\documentclass[preprints,article,accept,moreauthors,pdftex,10pt,a4paper]{mdpi} 
\usepackage{amsmath,amssymb,amsfonts}
\usepackage{graphicx}
\usepackage{textcomp}
\usepackage{amsmath}
\usepackage{algorithm}
\usepackage{float}
\usepackage[noend]{algpseudocode}
\usepackage{booktabs}
\firstpage{1} 
\pubvolume{xx}
\issuenum{1}
\articlenumber{5}
\pubyear{2019}
\copyrightyear{2019}
\history{Received: date; Accepted: date; Published: date}

\Title{Unsupervised Domain Adaptation using Generative Adversarial Networks for Semantic Segmentation of Aerial Images}

\Author{Bilel Benjdira $^{1,4,}$*, Yakoub Bazi $^{2}$, Anis Koubaa $^{3}$ and Kais Ouni$^{4}$}

\AuthorNames{Firstname Lastname, Firstname Lastname, Firstname Lastname and Firstname Lastname}

\address{%
$^{1}$ \quad Prince Sultan University, Saudi Arabia; bbenjdira@psu.edu.sa\\
$^{2}$ \quad Computer Engineering Department, College of Computer and Information Sciences, King Saud University, Riyadh 11543, Saudi Arabia; ybazi@ksu.edu.sa \\
$^{3}$ \quad Prince Sultan University, Saudi Arabia/Gaitech Robotics, China/CISTER, INESC-TEC, ISEP, Polytechnic Institute of Porto, Portugal; akoubaa@psu.edu.sa \\ 
$^{4}$ \quad Research Laboratory Smart Electricity \& ICT, SEICT, LR18ES44. National Engineering School of Carthage, University of Carthage, Tunisia; kais.ouni@esti.rnu.tn }

\corres{Author to whom correspondence should be addressed, email: bbenjdira@psu.edu.sa}

\abstract{Segmenting aerial images is being of great potential in surveillance and scene understanding of urban areas. It provides a mean for automatic reporting of the different events that happen in inhabited areas. This remarkably promotes public safety and traffic management applications. After the wide adoption of convolutional neural networks methods, the accuracy of semantic segmentation algorithms could easily surpass 80\%  if a robust dataset is provided. Despite this success, the deployment of a pre-trained segmentation model to survey a new city that is not included in the training set significantly decreases the accuracy. This is due to the domain shift between the source dataset on which the model is trained and the new target domain of the new city images. In this paper, we address this issue and consider the challenge of domain adaptation in semantic segmentation of aerial images. We design an algorithm that reduces the domain shift impact using Generative Adversarial Networks (GANs). In the experiments, we test the proposed methodology on the International Society for Photogrammetry and Remote Sensing (ISPRS) semantic segmentation dataset and  found that our method improves the overall accuracy from 35\% to 52\% when passing from Potsdam domain (considered as source domain) to Vaihingen domain (considered as target domain). In addition, the method allows to recover efficiently the inverted classes due to sensor variation. In particular, it improves the average segmentation accuracy of the inverted classes due to sensor variation from 14\% to 61\%.}


\begin{document}


\vspace{6pt} 
\section{INTRODUCTION}

Semantic segmentation is an image analysis task that assigns for every pixel in an input image a label that describes the class of its enclosing region. Beyond image classification  and object detection, semantic segmentation is the highest-level image analysis task that allows a complete scene understanding of the whole input image. \par\
Semantic segmentation was referred in many remote sensing works as pixel-wise classification. Semantic segmentation term is more used in computer vision and it is being more and more adopted in remote sensing.
Semantic segmentation can be used in aerial imagery in a variety of potential applications, like urban area monitoring and planning , traffic management and analysis, hazard detection and avoidance and so on. This potential is boosted by the increasing adoption of Unmanned Aerial Vehicles (UAVs). UAVs make the surveillance of inhabited areas easier due to their flexibility, great mobility and the high resolution images that they can gather and stream in real time. These images can be automatically processed by accurate semantic segmentation algorithms to reinforce substantially the ability to analyze and describe the surveyed scenes automatically. \par
\begin{figure*}
\begin{center}  
\includegraphics[width=14cm]{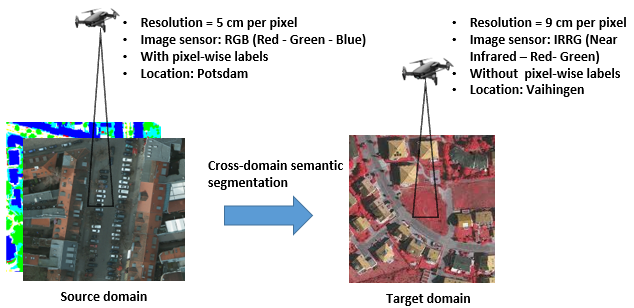}
\caption{\small \sl Cross-domain semantic segmentation in aerial imagery
\label{fig:cross_domain_sem}}  
\end{center}  
\end{figure*}
The progress of semantic segmentation algorithms was delayed years ago by the low accuracy of traditional approaches of image analysis algorithms based on the extraction of hand-crafted features. But since the emergence of highly descriptive feature extractors like convolutional neural networks, the whole area of image analysis has shown a significant increase of accuracy. In fact, since the year 2012\cite{krizhevsky2012imagenet}, convolutional neural networks (CNNs) have shown an outstanding efficiency in computer vision. Although the CNN architecture was first introduced in 1998, the adoption of CNN was inhibited by the heavy computations needed to train it for normal computer vision tasks. However, this issue  has recently been resolved by the improvements made to the Graphic Processing Units (GPUs). These improvements made possible the parallelization of the training computations over the multiple cores of the GPU, thus reducing the training time for CNNs and promoting the adoption of such a model in real image analysis tasks. This success may be explained by the fact that CNNs can extract multiple levels of  representation for the objects of interest. These representations are trained through the different layers to be lighting-invariant, scale-invariant and rotation-invariant to extract the object of interest efficiently. 

This advancement enhanced the areas of semantic segmentation algorithms. Recently, several CNN- based architectures have shown their efficiency in this task, such as fully connected network (FCN)\cite{long2015fully}, SegNet\cite{Badrinarayanan_2017}, UNet\cite{ronneberger2015u}, PSPNet\cite{Zhao_2017} and  DeepLab\cite{chen2018deeplab}. If a robust dataset is provided and semantically labeled, training one of the state of the art models could lead easily to an accuracy that exceeds 80\%\cite{lateef2019survey}. 

Despite this notable success made in the area of semantic segmentation algorithms, a great challenge is hampering their implementation in real use cases. In fact, if we train a model on a specific dataset, the accuracy will be high when applying this model on images belonging to the same domain of the train set (lighting conditions, sensor type, resolution, object representation). But if we try to apply this model to segment images acquired under different conditions, the performance falls dramatically due to the domain shift between the images used in the source domain (used during the training) training and the target domain. To illustrate this fact, we made an experiment where we chose a state of the art segmentation algorithm (DeepLab v3 plus\cite{Chen_2018}) which is trained on International Society for Photogrammetry and Remote Sensing (ISPRS) Potsdam benchmark dataset\cite{gerke2014use} and we applied it  for segmenting a random image form ISPRS Vaihingen benchmark dataset. A drop in global accuracy from 85\% to 35\% was observed (see experiments description in section IV for more details). \(Figure\) \ref{fig:cross_domain_sem} shows a typical situation in which we have a trained model on a specific source domain and we want to use this model to segment another domain. The domains have different characteristics (resolution per pixel changed from 5 centimeters to 9 centimeters, image information are changed from Red-Green-Blue sensor to Near Infra Red- Red-Green, location is changed from Potsdam to Vaihingen). 
The ordinary solution to cope with this intriguing limitation is to make a new semantically labeled dataset on the target domain and to train the model on it. This solution is very costly and impractical. In fact, collecting a large dataset of pixel-labeled images for the targeted city of interest will be time-consuming and expensive. Indeed, pixel-labeling of Cityscapes images (size is 2040 by 1016 pixels) takes 90 minutes on average\cite{cordts2016cityscapes}. Remote sensing are more time-demanding as they contains objects from different sizes (small sized objects like cars and roads  need more attention and effort in the labeling process). To reduce human efforts in manual pixel-wise classification, a number of solutions have been introduced like synthesizing data from 3D rendered images \cite{richter2016playing, ros2016synthia} or weakly supervised labeling\cite{saleh2016built,bearman2016s, shimoda2016distinct}. However, these approaches still have limitations as they also  require significant human efforts. Moreover, they have some drawbacks (like domain shift from 3D rendered images to real images in synthetic data solutions and imprecise boundaries in weakly supervised solutions). This is why it is highly fruitful to invest for an automatic domain adaptation solution. 

Domain adaptation is the machine learning field that aim at learning from a source data distribution how to improve the performance of a model on a different target data distribution.It addresses to reduce the domain shift problem between the source domain dataset used in training and the target domain dataset. For this purpose, we typically design a mapping function between the source domain data and the target domain data. Recent domain adaptation techniques use deep learning models for training this mapping function\cite{tzeng2015simultaneous, long2015learning, tzeng2017adversarial,luo2017label}. Domain adaptation techniques could also consolidate this mapping function by adding some modifications on the model itself to get correlated feature level with the target domain dataset. 

Inspired by recent advances in Generative Adversarial Networks (GANs) \cite{goodfellow2014generative, Zhu_2017}, we developed an algorithm for domain adaptation for aerial imagery based on GANs. The objective of our method is to handle the scenario presented in \(Figure\) \ref{fig:cross_domain_sem} and similar cases. We aim to add the ability for a semantic segmentation model to handle domains that are different from the source domain with minimal costs and maximum accuracy. Our method is divided into two steps. The first step considers the process of converting the images of the dataset from the source domain to the target domain. This is done using a GAN model trained using a cyclic-loss to map between two sets, one is taken from the source domain and the other is taken from target domain. We adopted this approach to eliminate the need for  paired set of images which may be time consuming. The second step is to fine-tune the already trained semantic model using the mapped version of the dataset associated with the original labels. After the fine-tuning process, the model will improve its ability to semantically label images taken form the target domain. 
The major contribution of our work can be presented as follows: (1) First,to the best of our knowledge, no previous works addressed the problem of domain adaptation for semantic segmentation in aerial imagery using GANs. (2) Second, we demonstrate that our approach mitigates the domain shift problem for cross-domain semantic segmentation in aerial imagery which allows the portability of the semantic segmentation model over different image domains. (3) Third, we validate the method on the ISPRS semantic labelling dataset by making cross-domain semantic segmentation between Potsdam dataset and Vaihingen dataset. (4) Fourth, we introduce GANs as promising solution for analysis of aerial imagery. 

The rest of the paper is organized as follows: Section 2 will make an overview about the related works in area of domain domain adpatation in semantic segmentation. Section 3 makes an introduction to GANs. Section 4 will describe our proposed method. Section 5 will present the experimental details we made to test our method. Section 6 will discuss its efficiency  for domain adaptation in aerial imagery. Section 7 concludes our work and deduces about the contribution we made in this paper.  

\section{Related works}
In this section, we will discuss the related works on domain adaptation in semantic segmentation.
\subsection{Domain adaptation}
When applying a machine learning algorithm, we generally assume that the training data and the test data are belonging to the same underlying distribution. But in real scenarios, we face some discordance between them. This discordance decreases the efficiency of the model  outside its training domain. Domain adaptation is a separate field in machine learning that aims to rectify this discordance and help the model to be better generalized to test domains. \par
The efforts on domain adaptation in image analysis are focusing on classification and regression tasks \cite{Patel2015VisualDA} like trying to train models on online photos to classify objects in real world\cite{saenko2010adapting}. Recent works are mostly oriented towards improving the adaptability of deep learning algorithms\cite{long2015learning, tzeng2015simultaneous, ganin2016domain, ganin2014unsupervised, bousmalis2017unsupervised}.
\subsection{Domain adaptation on semantic segmentation}
Many works on this field are using simulated data \cite{hoffman2017cycada,ros2016synthia, vazquez2014virtual, peng2018synthetic, shrivastava2017learning, shafaei2016play}. In fact, they expected to use domain adaptation to improve the segmentation efficiency on real images by training models on synthetic data. Among the first works that treated domain domain adaptation on semantic segmentation, we can find FCNs in the wild\cite{hoffman2016fcns} who employed a pixel-level adversarial loss to guide the model towards learning the domain-invariant features. The goal is to make the adversarial classifier not to differentiate between source and target domains to equalize his performance on both domains. Hoffman et al. proposed CyCADA\cite{hoffman2017cycada} as another method that convert the source images (synthetic data ) to the style of the target (real datat) using CycleGAN. The converted images are then feeded to the segmentation model to improve his performance on the target images. 
Zhang et al. \cite{zhang2017curriculum} proposed  a curriculum-style learning approach to minimize the domain shift. They concludes properties of the target data by combining the learning of the local distributions over landmark superpixels with the learning of global label distribution. Then they trained the segmentation network by regularizing it to follow those concluded properties. Chen et al. \cite{chen2018road} proposed ROAD (Reality Oriented Adaptation) by designing two losses to align the source and the target domains. The first is called target guided distillation loss and the second is a spatial aware adaptation loss. We divide the feature map of the image into grids. Then we calculate a maximum mean discrepancy loss for every grid. Sankaranarayanan et al.\cite{sankaranarayanan2017unsupervised} proposed an auto-encoder network that takes as input both source and target images and regenerate them before being feeded to the segmentation network. Tsai et al.\cite{tsai2018learning} proposed CGAN to add random noise to the source data before being feeded to the segmentation network. They found that this approach improve the performance of the model on target domains. Huang et al. \cite{huang2018domain} trained separately two models for the source and the target domains. Because the target domain is without labels, the target model is trained by regressing it to the weights of the source model. Also, an adversarial loss is calculated in every layer of the two networks. Zhang et al.\cite{Zhang2018FullyCA} used an adversarial loss between the source and the target data on both the first layers and the layers of the network. This method improves the adaptation performance of the network. \par

These are the main works that treated domain adaptation on semantic segmentation. We can deduce that, up to our knowledge, no one treated domain adaptation on semantic segmentation on aerial imagery. Most of the methods treated images of urban scenes taken from a camera mounted on a car. Aerial imagery has many dissimilarities with the data treated in these works. This is why we targeted this problem in this paper. We used to test the efficiency of our method on the International Society for Photogrammetry and Remote Sensing (ISPRS) semantic segmentation dataset. We studied the domain adaptation from Potsdam domain dataset to Vaihingen dataset\cite{gerke2014use}. 

\section{Generative Adversarial Networks (GANs)}
\subsection{Generator and Discriminator}
\begin{figure}[H]  
\begin{center}  
\includegraphics[width=8cm]{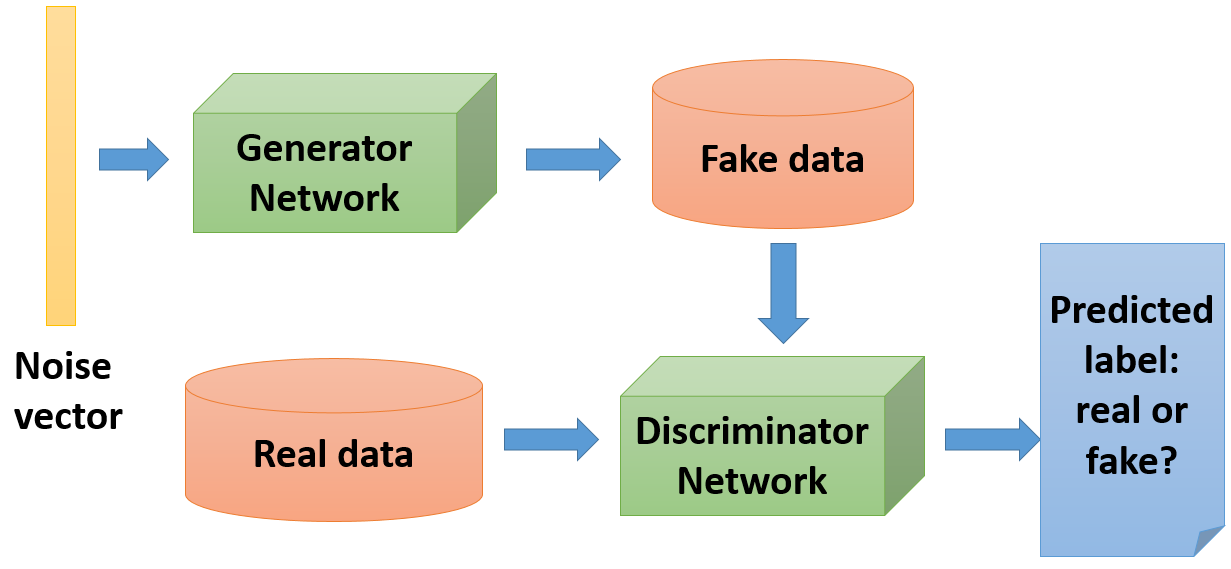}
\caption{\small \sl Generative Adversarial Network\label{fig:GAN}}  
\end{center}  
\end{figure}

GANs are increasingly being popular due to the wide area of applications that they address. It was firstly introduced in 2014 by Ian Goodfellow et al.\cite{goodfellow2014generative}. It is composed of two models named respectively generator and discriminator. The generator model is trained to generate data that are similar to the real data considered. The discriminator is trained to differentiate between the real and fake data generated by the generator. During the training, the generator and the discriminator are competing with each other, playing an adversarial zero-sum game. The loss on both models is balanced by the loss of its adverse model. In fact, the generator is trained to generate fake data that fool the  discriminator making it judge the generated fake data as real data. On the other side, the discriminator is trained to differentiate between the fake data and the real data. During the training, this game is solved using game theory theorems. At the end, the generator is well trained to generate data that are similar to the real data and not previously seen in the training set. The discriminator is well trained to differentiate between the real and fake data. This simultaneous training of discriminator and generator is drawn in the \(Figure\) \ref{fig:GAN}. 
The two networks compete with each other during the training until reaching the Nash equilibrium. In game theory, Nash equilibrium is a strategy profile in which no player can unilaterally deviate and improve his payoff\cite{oliehoek2018local}.

\par 
The GAN's objective function is described by \(Equation\) \ref{eq:1}:
\begin{multline} \label{eq:1}
\begin{gathered}
min_{G}max_{G}V(D,G) =  {\rm I\!E}_{X \sim P_{data}(X)}[logD(X)] +  {\rm I\!E}_{z \sim P_{z}(z)}[log(1-D(G(z)))]
\end{gathered}
\end{multline}

where \(G\) is the generator trained by maximizing \(D(G(z))\). \(D\) is the discriminator trained by minimizing \(D(G(z)\). \(X\) is an image sampled from the real data distribution \(p_{data}\), \(z\) is the noise vector sampled from the distribution \(p_z\), \(G(z)\) is the fake image generated by the generator. \(D\) and \(G\)  are playing the two-player minimax game with value function \(V (G, D)\)\cite{goodfellow2014generative}. \par

GANs have a plethora of implementations and applications\cite{Goodfellow2016NIPS2T}. The most attractive application  that we may get inspired for domain adaptation is Image to Image translation. In the next subsection, we will focus more on this area and introduce the GAN models designed for this task. 

\subsection{GAN for Image to Image translation}
Image to image translation is the task of converting one image from a domain to another. For example,  translating an image taken in summer to another one that mimic its appearance if it were taken in winter. This area may have numerous applications and use cases and many GAN models were designed in literature\cite{Liu2017UnsupervisedIT, Isola2017ImagetoImageTW,Zhu2017TowardMI, Yi2017DualGANUD}. Image translation can be either paired\cite{pix2pix_2017} or unpaired\cite{Zhu_2017}. 
\subsubsection{Paired Image translation}
In paired image translation, the GAN model should be trained in a supervised way using labeled pairs from source domain to target domain. Considering \(X\) is the source dataset, \(Y\) is the target dataset and \(N\) is the number of samples in every dataset, the model will access every pair of corresponding images\(\{x_i,y_i\}_{i=0..N}\) and try to learn how to convert between \(X\) and \(Y\) domains based on these samples. Pix2pix\cite{pix2pix_2017} is the major state of the art architecture for paired image to image translation.  
\subsubsection{Unpaired Image translation}
In unpaired image translation, the GAN model is trained in a unsupervised way between two sets of images. The first set represents the source while the second represents the target. Considering \(X\) is the source dataset, \(Y\) is the target dataset and \(N\) is the number of samples in every dataset, \(\{x_i\}_{i=0..N}\) and \(\{y_i\}_{i=0..N}\) are not necessarily corresponding and could be taken randomly from the associated domain set. CycleGAN\cite{Zhu_2017} is the major state of the art architecture for unpaired image to image translation. It make a bidirectional image to image translation between two sets of images.
\par



\section{Proposed method}
\subsection{Our proposed GAN architecture}
\begin{figure*} 
\begin{center}
\includegraphics[width=14cm]{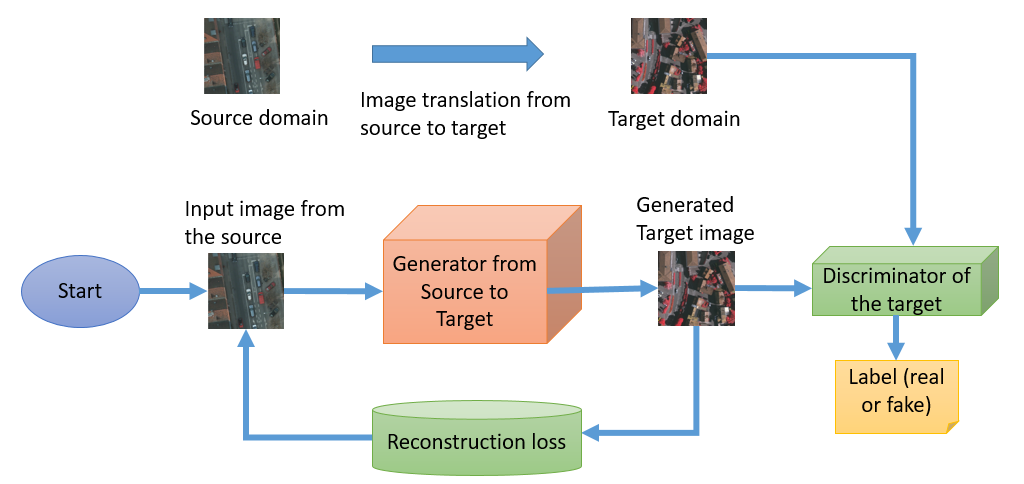}
\caption{\small \sl GAN architecture for unpaired image translation in aerial images \label{fig:cyclic-gan}}  
\end{center}  
\end{figure*}

The proposed method aims to perform image level translation from the source domain to the target domain using a GAN as shown in \(Figure\) \ref{fig:cyclic-gan}. We described in this figure how we implemented an unpaired image to image translation GAN from the source domain to the target domain. 
 \par 
This procedure was designed to make images of  the source domain mimic the characteristic of the target domain (types of sensors, quality of the images, resolution...). This will have the effect to reduce the domain shift related to the quality and characteristics of the images in the training set. To reduce our method cost, we did not adopt the traditional GAN approach. In fact, if we adopt it without modification, a paired dataset should be provided for every class of objects considered in our model. This will be really costly and time-demanding and do not harmonize with our goal to make the domain adaptation straightforward and easy to implement. Hence, we adopted a modified approach inspired from many state of the art architectures\cite{Zhu_2017, rs10020351}. We implemented an unpaired  image translation adversarial network working  in  unidirectional way from the source to the target as shown in the \(Figure\) \ref{fig:cyclic-gan}. The translation of an image from the source domain to the target domain does not need paired images. Images for both domains are collected separately without the need for corresponding pairs to train a mapping function \(G:X\rightarrow Y\). This function \(G(X)\) learns during the training process to make images from the source \(X\) imitate the distribution of images in the target \(Y\) minimizing an adversarial loss. But, we have here to take in consideration another condition. If we will only be limited to this mapping function, the image translation will not be done as expected. In fact, because this mapping function is not constrained with paired data, the image translation is prone to be done in a meaningless way leading to a model collapse. Therefore, we consider to add the inverse mapping function \(F:Y\rightarrow X\) that makes the image translation on the inverse direction from the target to the source. This function \(F(Y)\) learns during the training process to imitate the distribution of images in \(X\) minimizing a second adversarial loss. Then, we add the reconstruction loss to consolidate that \(F(G(X))\approx   X\) and \(G(F(X))\approx X\) simultaneously. Then we train our model jointly so that the image structure will be conserved during the translation process from the source domain to the target domain. \par
\begin{figure}[H]
\begin{center}
\includegraphics[width=5.5cm]{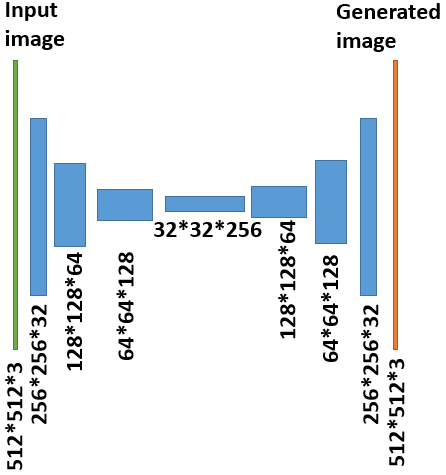}
\caption{\small \sl The Encoder-Decoder architecture of the Generator \label{fig:generator_architecture}}  
\end{center}  
\end{figure}

The architecture of the generator is similar to U-Net\cite{ronneberger2015u}  architecture. We used an encoder decoder network as illustrated in \(Figure\) \ref{fig:generator_architecture}. Four convolutional layers are set for Downsampling and four convolutional layers are used for upsampling.  We used Leaky ReLU\cite{leaky_relu} as the activation function for all the layers of downsampling and Standard ReLU for all the layers of upsampling. Leaky ReLU is similar to the standard ReLU (Rectified Linear Unit) but has a small slope \(\alpha\) in the negative region. The Leaky ReLU function is defined as \(f(x) = x if x>=0\) and as \(f(x) = \alpha  if x <0\), where \(\alpha\) is a very small coefficient. It allows to have small positive gradient when the function is not activated. The output features extracted from the encoder are passed into the decoder that will learn how to rebuild the original feature vector. We used dropout\cite{Srivastava2014DropoutAS} in the decoder architecture to reduce overfitting. We used Instance Normalization \cite{ulyanov2016instance} after every layer in the generator, because it is proven in \cite{ulyanov2016instance} that it works better than batch normalization\cite{ioffe2015batch} for generator neural networks. It helps to provide better  stylization in the image generation process. \(Figure\) \ref{fig:generator_architecture} shows the architecture of the generator. \par
 \par
\begin{figure}[H]
\begin{center}
\includegraphics[width=7cm]{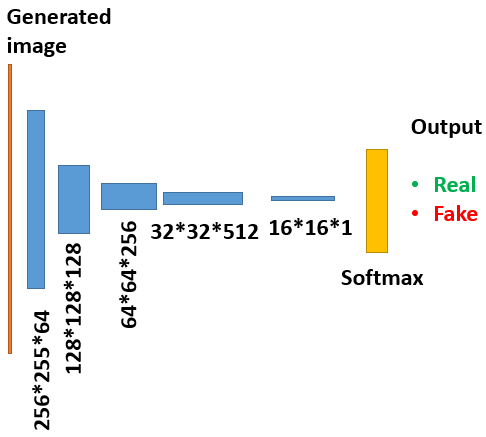}
\caption{\small \sl The architecture of the discriminator \label{fig:discriminator_architecture}}  
\end{center}  
\end{figure}

Concerning the discriminator architecture, it receives as input the generated image and makes a binary classification output to real or fake image. We used five convolutional layers that encodes the generated image into a feature vector of size 256. Then we use the softmax function to convert this feature vector into a binary output. In the same way as the generator, we used the Leaky ReLU \cite{leaky_relu} as  activation function for all the layers of the network and we applied Instance Normalization\cite{ulyanov2016instance} in every layer of the discriminator except the first layer. \(Figure\) \ref{fig:discriminator_architecture} shows the architecture of the discriminator. 
 \par
\subsection{Algorithm description}
\begin{figure*} 
\begin{center}
\includegraphics[width=14cm]{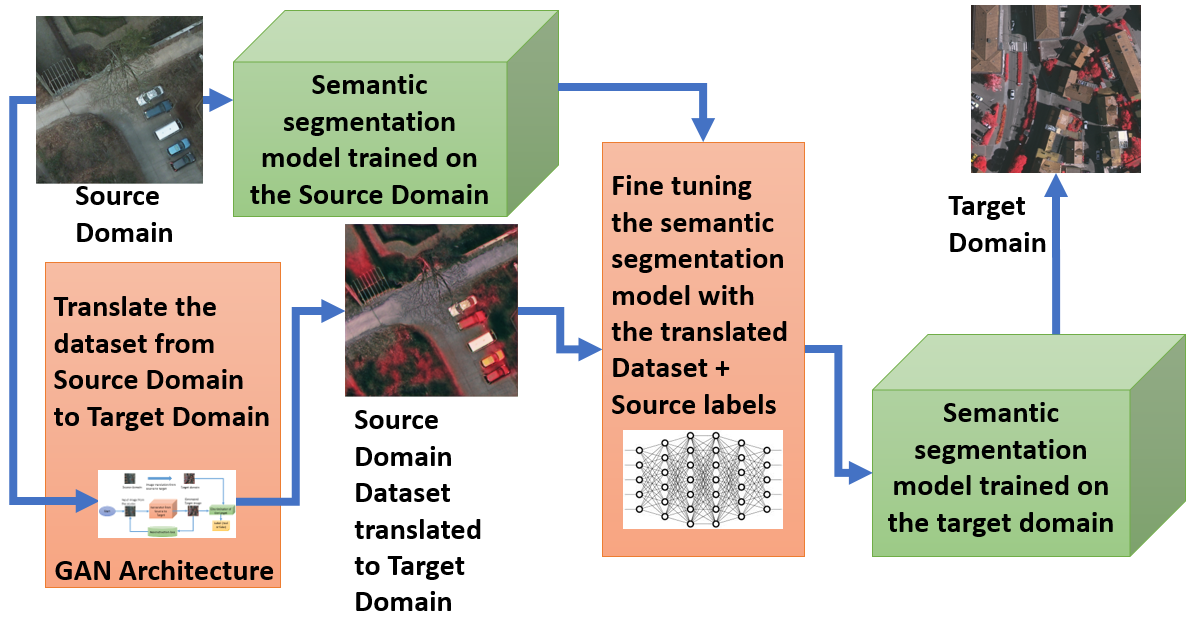}
\caption{\small \sl Flowchart of the domain adaptation algorithm \label{fig:proposed-methodology}}
\end{center}  
\end{figure*}

Based on the GAN architecture provided in the \(Figure\) \ref{fig:cyclic-gan}. We designed and implemented our proposed algorithm for domain adaptation in aerial imagery. The flowchart of the algorithm is described in the \(Figure\) \ref{fig:proposed-methodology}. 
The algorithm is divided into four steps. The first step is to train a segmentation model on the source dataset. In principle, with a good structured dataset, the segmentation accuracy could reach easily a level higher than 80\%. The second step considers the training of our proposed GAN architecture to translate image efficiently from the source domain to the target domain. The third step is to convert the source dataset to the target domain using this GAN architecture.The output of the third step  is new dataset that conserves the structures represented in the images of the source dataset but mimics the global characteristics of the target dataset (imaging sensors, global coloring...).  The fourth step is to fine-tune the already trained segmentation model with the translated dataset associated with the source labels. This step helps the model parameters to learn the patterns of the target dataset and to converge to a better recognition of image structure on the target dataset. After the fine-tuning process, the semantic segmentation model is adapted to work on the target dataset. 
 \par

\subsection{Problem formulation}
In this section, we present the formal mathematical model of the algorithm. We consider the problem of domain adaptation from a source domain data \(X{_S}\), that are already mapped to their labels \(Y{_S}\), to a target domain data  \(X{_T}\) without labels. \par
We start by training a source model \(M{_S}\) that performs the semantic segmentation on the source data by mapping the input images and their corresponding labels. The pixel-wise labels  have one of \(C\) classes. Using the cross-entropy loss function, the source model  \(M{_S}\) corresponds to:\par
\begin{multline} \label{eq:2}
\begin{gathered}
L_{M_S} (M_S, X_S, Y_S) = - {\rm I\!E}_{(x_s,y_s) \sim (X_S, Y_S)}  \sum_{c=1}^C {\rm I\!l}_{[c=y_s]}log( Softmax (M^{(c)}_S (x_s)))
\end{gathered}
\end{multline}
Thanks to the advance in the semantic segmentation algorithms, \(M_S\) generally performs well on the source data. However, when applying the source model \(M_S\) on the target data we will have lower accuracy due to the domain shift that exists between the source and the target domain. To alleviate this domain shift, we will begin first by mapping the dataset images of the source domain to the target domain. This is implemented by our proposed GAN architecture that learns how to map the image samples between domains so that the discriminator will be unable to detect that the mapped image from the source to the target does not belong really to the target. The next step is to fine-tune the source model \(M{_S}\) by running the trained model on the mapped dataset and this helps to generalize our source model to perform better on the target domain as proven in the experimental section of this paper. \par
The mapping model from source to target \(G_{S \longrightarrow T}\) is implemented and trained to map from the source domain to the target domain. The goal is to generate image samples that will be classified by the adversarial discriminator \(D_T\) as real images from the target domain. On the other side, the adversarial discriminator \(D_T\) is trained to not be fooled by the generated images and to detect them successfully as fake. The loss function corresponding to this is: \par
\begin{multline} \label{eq:3}
\begin{gathered}
L_{GAN} (G_{S \longrightarrow T}, D_T, X_T, X_S) = {\rm I\!E}_{x_t \sim X_T}[logD_T(x_t)] +  {\rm I\!E}_{x_s \sim X_S}[log(1-D_T(G_{S \longrightarrow T}(x_s)))]
\end{gathered}
\end{multline}

The training of this loss makes \(G_{S \longrightarrow T}\) capable of generating from a sample image taken from the source domain an image that imitates the appearance of an image taken from the target domain. Therefore, from the source segmentation model \(M{_S}\), we make a new model \(M{_T}\) that minimizes the loss function:\par
\begin{multline} \label{eq:4}
L (M_T, G_{S \longrightarrow T}(X_S), Y_S) = \\
- {\rm I\!E}_{(G_{S \longrightarrow T}(x_s),y_s) \sim (G_{S \longrightarrow T}(X_S), Y_S)}  \sum_{c=1}^C {\rm I\!l}_{[c=y_s]}log( Softmax (M^{(c)}_T (G_{S \longrightarrow T}(x_s))))
\end{multline}\par
This loss function is trained in a similar manner as the loss defined in \(Equation\) \ref{eq:3}. 
Therefore, the target model \(M{_T}\) is a copy  from the already trained source model  \(M{_S}\) that we train on the mapped dataset by minimizing the loss defined in the \(Equation\) \ref{eq:4}. This operation makes the model generalized better on the target domain. The GAN loss defined in the \(Equation\)  \ref{eq:4} ensures that for a sample image \(x_s\) from the source domain, \(G_{S\longrightarrow T}(x_s)\) will resemble the sample images taken from the domain \(X_T\). Although general resemblance can be assured through the training, we cannot guarantee that \(G_{S\longrightarrow T}(x_s)\) maintains the structural content of  \(x_s\). \par
To preserve the content and the structure of \(x_s\) during the mapping operation assured by
\(G_{S\longrightarrow T}\), we a GAN network working on the inverse direction from the target to the source as detailed in the section III.A. It maps from the target to the source \(G_{T\longrightarrow S}\). The loss to train for \(G_{T\longrightarrow S}\) is identical to the loss defined for \(G_{S\longrightarrow T}\) in \(Equation\) \ref{eq:3}, just the parameters of the loss are changed to be as: \par
\begin{equation} \label{eq:5}
\begin{gathered}
L_{GAN} (G_{T \longrightarrow S}, D_S, X_S, X_T)
\end{gathered}
\end{equation}\par
Then, we ensure that mapping a sample image \(x_s\) from the source to the target using \(G_{S\longrightarrow T}\) followed by another mapping of this generated image \(G_{S\longrightarrow T} (x_s)\) back to the source using the mapping function \(G_{T\longrightarrow S}\) will generate an identical image of the source \(x_s\). This is the reconstruction loss constraint that we added as we explained in III.A to keep the structural content of the images during the mapping process. This loss constraint is formulated by the following equations:
\begin{equation} \label{eq:6}
G_{T \longrightarrow S} (G_{S \longrightarrow T } (x_s)) \approx x_s
\end{equation}
\begin{equation} \label{eq:7}
G_{S \longrightarrow T} (G_{T \longrightarrow S }(x_t))  \approx x_t
\end{equation}\par
To ensure that the \(Equation\) \ref{eq:6} and \ref{eq:7} are satisfied, we impose the reconstruction loss constraint defined in the following equation:
\begin{multline} \label{eq:8}
L_{reconstruction} (G_{S \longrightarrow T}, G_{T \longrightarrow S}, X_S, X_T) = \\
{\rm I\!E}_{x_s \sim X_S}[\Vert G_{T \longrightarrow S} (G_{S \longrightarrow T } (x_s)) - x_s \Vert_1]  
 + {\rm I\!E}_{x_t \sim X_T}[\Vert G_{S \longrightarrow T} (G_{T \longrightarrow S } (x_t)) - x_t \Vert_1] 
\end{multline}
After finishing the training of our proposed GAN architecture, we use it to translate the source data  \(X_S\) to \(X_{S\_tr}\).Then, we profit from the labels provided with the source data by reusing theme exactly the same in the training with the new translated dataset. We take the segmentation model \(M_S\) which is already trained on source data before translation, we fix the weight values and use it as a start point for the training of our target model \(M_T\). This model performs the semantic segmentation on the translated image data by mapping \(X_{S\_tr}\) with their corresponding pixel-wise labels  \(Y{_S}\). Using cross-entropy as loss function, the target model corresponds to:\par
\begin{multline} \label{eq:9}
\begin{gathered}
L_{M_T} (M_T, X_{S\_tr}, Y_S) = 
- {\rm I\!E}_{(x_{s\_tr},y_s) \sim (X_{S\_tr}, Y_S)} \sum_{c=1}^C {\rm I\!l}_{[c=y_s]}log( Softmax (M^{(c)}_T (x_{s\_tr})))
\end{gathered}
\end{multline}
Finally, we obtain a target model \(M_T\) that it is more adapted to work on the target domain as described in the Experimental section.

\section{Experimental results}
In this section, our objective is to prove the efficiency of the proposed algorithm by providing the description of the implemented experiments and discussing the obtained results. 
\subsection{Datasets and evaluation metrics}
\subsubsection{Datasets}
\begin{figure}  
\begin{center}  
\includegraphics[width=7cm]{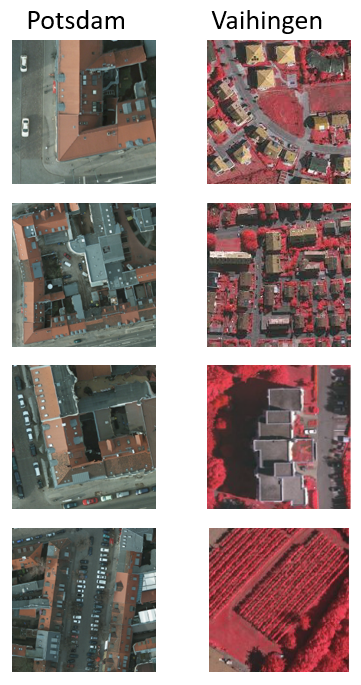}
\caption{\small \sl Samples of images from Potsdam and Vaihingen ISPRS datasets
\label{fig:samples_dataset}}  
\end{center}  
\end{figure}

To validate our methodology, we used the ISPRS (WGII/4) 2D Semantic segmentation benchmark dataset\cite{gerke2014use}. It is afforded by the ISPRS 2D Semantic Labeling Challenge that provides currently, the best platform to evaluate semantic segmentation algorithms for aerial images. We used respectively the Vaihingen and Potsdam datasets which are publicly available to the community. Although Digital Surface Model (DSM) data is provided for every image, we only use the image data as we are targeting domain adaptation using only image data. Both datasets contain very-high resolution images with a resolution of 9 cm for Vaihingen images and 5 cm for Potsdam images. Note that the resolutions are different in both datasets and this represents one of the factors that requires domain adaptation. These resolutions are categorized in aerial imagery as Very High Resolution (VHR) and it is helpful in  recognizing objects clearly. Besides, this helps to maximize the intra-class variance and minimize the inter-class variance by providing more details about objects. All images in both datasets are provided with their semantic segmentation labels which comprises six classes of ground objects: building, tree, car, impervious surfaces, low vegetation and clutter/background. The impervious surfaces indicate paved area with no building on it. The clutter/background category refers to all the ground objects that are not included in the other five categories. The Vaihingen dataset includes 33 TOP images with sizes near to  \(2000*2000\) pixels. All these 33 TOP images are released with the ground truth. The TOP file contains three channels:  Infrared, Red and Green bands. Among the 33 TOP images, 27 TOP images are used for training and 6 images are used for the test. The Potsdam dataset is a larger dataset that contains 38 TOP images with a fixed size of \(6000*6000\) pixels. All these images are released with their ground truth. The TOP files for Potsdam contains 4 different spectral channels: Red, Green, Blue and InfraRed. Among the 38 TOP images, 32 images are used for the training and 6 images are used for the test. To train the segmentation model, we divided the images and their labels into squares of size \(512*512\) and feed the network with uniform patches of size \(512*512\). \(Figure\) \ref{fig:samples_dataset} shows samples from Potsdam and Vaihingen ISPRS datasets. 

The distribution of pixels over the six classes is not proportionally balanced. Categories like Impervious Surface or Buildings are much more represented as compared to other classes like Cars or Clutter. \(Table\) \ref{tab:table_dataset} represents the percentage of each class proportionally to the total number of pixels. The percentage of a class is calculated by summing the number of pixels belonging to this class divided by the total number of pixels in the dataset. 

\begin{table}[]
\centering
\begin{tabular}{@{}lll@{}}
\toprule
Category & Potsdam & Vaihingen \\ \midrule
Impervious Surfaces & 29.9\% & 29.3\% \\ 
Buildings & 28.2\% & 26.9\% \\ 
Low vegetation & 20.9\% & 19.4\% \\ 
Trees & 14.4\% & 22.4\% \\ 
Cars & 1.7\% & 1.3\% \\ 
Clutter & 4.8\% & 0.7\%  \\ \bottomrule
\end{tabular}
\caption{\small \sl Percentage of each category in the datasets}
\label{tab:table_dataset}
\end{table}
\subsubsection{Domain shift analysis}
The domain shift from the source domain (Potsdam) to the target domain (Vaihingen) is resulted from 3 essential factors. The first factor is the imaging sensor factor. Images of Vaihingen are captured using a sensor IRRG ( Infrared, Red, Green). The images of Potsdam are captured using a 4 bands sensor RGBIR (Red, Green, Blue, Infrared). For example, the class vegetation and trees are characterized by the green color due to the sensor RGBIR used for Potsdam dataset. The segmentation model will be trained to recognize the varieties of green color that helps to identify these classes accurately. In Postdam dataset, the green color is well represented. In Vaihingen dataset, it is totally transformed to a red color due to the change of the sensor. This change will affect the accuracy of the segmentation model and make a significant domain shift. The second factor is the resolution factor. Images of Vaihingen are captured using a resolution of 9 centimeters per pixel. Images of Potsdam are captured using a resolution of 5 centimeters per pixel. Going from one resolution to another could affect the ability of the segmentation model to identify accurately the classes and therefore generate a domain shift. The third factor of domain shift is the structural representation of the classes. Many classes show a difference of representation passing from  Potsdam dataset to Vaihingen dataset. For example, Buildings in Postdam and Vaihingen are very comparable as they correspond to the building style of modern German towns. There is not much difference in the representation of the class Building when going from Potsdam to Vaihingen. But for other classes like Low vegetation and Trees, there is a clear difference. In fact, Vaihingen contains agricultural areas while Potsdam does not contain this kind of areas. Types of trees and vegetations differs  when switching between the two datasets. The difference is  clearer in the low vegetation class than the trees class. In fact, there are similarities between most tree types of Vaihingen and Potsdam. 
\par
The domain shift between Potsdam and Vaihingen is generated  from a combination of the three factors(imaging sensors, resolution, class representation). This allows us to study the effect of our proposed algorithm on reducing the domain shift related to every factor. \(Table\) \ref{tab:domain_shift_factors} summarizes the effect of these factors on the domain shift of every class. The estimation of the factor impact is made after a careful analysis of every class on both domains. We can note that the effect of the resolution on the domain shift is low on all classes. In fact, passing from 5 centimeters per pixel to 9 centimeters per pixel does not affect very much the accuracy of the segmentation model. The feature extraction layers of the model have the ability to manage this scale of resizing. We note that the class Building is mostly affected by the sensor factor, thus it will be a study case for the effect of our algorithm on reducing domain shift made by the sensor factor only. The Trees class will be similarly a study case as it is mostly affected by the sensor factory and moderately affected by the class representation factor. Impervious surfaces and cars classes are not really affected by the three factors, they will be a study case for the effect of our algorithm on classes that are not subjected to a domain shift when passing from one dataset to another. Classes Low vegetation and clutter are highly affected by the sensor factor and the class representation factor. They will be a study case to study the effect of our algorithm on reducing the domain shift related to these factors combined. 
\begin{table}[]
\centering
\begin{tabular}{@{}llll@{}}
\toprule
Factor of domain shift & Resolution & Sensor & Class Representation \\ \midrule
Impervious Surfaces & low & low & low \\ 
Buildings & low & high & low \\ 
Low vegetation & low & high & high \\ 
Trees & low & high & medium \\ 
Cars & low & low & low \\ 
Clutter & low & high & high \\ \bottomrule
\end{tabular}
\caption{\small \sl Effect of the domain shift factors on every class when passing from Potsdam dataset to Vaihingen dataset}
\label{tab:domain_shift_factors}
\end{table}

\subsubsection{Evaluation metrics}
To measure the efficiency of the semantic segmentation algorithms, we used four measures: the accuracy, the precision,  the recall and the F1 score. They are expressed using \(TP\) (True Positives), \(TN\) (True Negatives), \(FP\) (False Positives) and \(FN\) (False Negatives). If we consider a class \(C\), \(TP\) corresponds to the number of pixels classified as \(C\). \(TN\) is the number of pixels that don't belong to the class \(C\) and the segmentation model did not classified them as \(C\). \(FP\) is the number of pixels that are classified falsely as \(C\) while they belong to other classes. \(FN\) is the number of pixels that belong to the class \(C\) but the segmentation model associated them falsely to other classes. These measure are defined below: 
\begin{equation} \label{me:1}
Accuracy = \frac{TP+TN}{TP+TN+FP+FN} 
\end{equation} 
\begin{equation} \label{me:2}
Precision = \frac{TP}{TP+FP} 
\end{equation} 
\begin{equation} \label{me:3}
Recall=Sensitivity= \frac{TP}{TP+FN}
\end{equation} 
\begin{equation} \label{me:4}
F1 Score= 2*\frac{Pecision*Recall}{ (Precision + Recall)} 
\end{equation}
Moreover, we used also the Intersection over Union (IoU) to measure the efficiency of the segmentation. Since we have six different classes, IoU is calculated for every class separately. Then, the Mean IoU of all classes is caculated.  \(Equation\) \ref{me:5} represents how to calculate the IoU for two different data samples \(A\) and \(B\).
\begin{equation} \label{me:5}
IoU (A,B) = \frac{A \cap B}{A\cup B}
\end{equation}

\subsection{Experimental settings}
\subsubsection{Step1: Training the segmentation model}
\begin{figure}[H] 
\begin{center}  
\includegraphics[width=8cm]{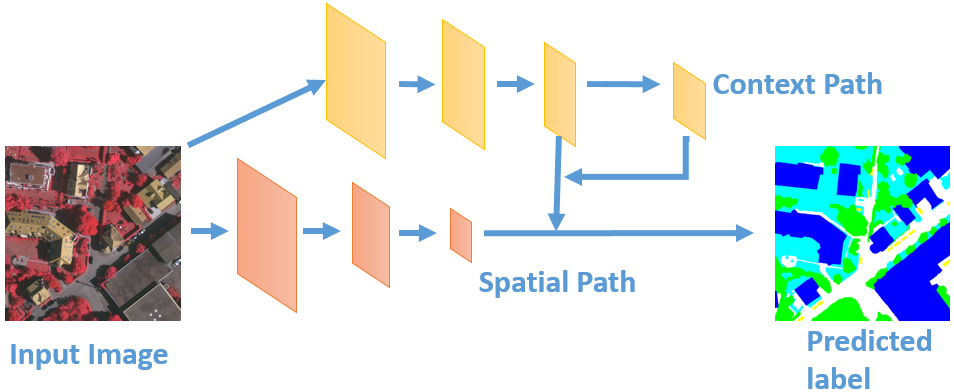}
\caption{\small \sl Architecture of Bilateral Segmentation Network (BiSeNet) \label{fig:bisenet}}  
\end{center}  
\end{figure}

We first start with training a segmentation model on the source dataset. We chose Potsdam as the source dataset because it is far greater than Vaihingen dataset. In fact, in real scenarios, target datasets are smaller and less structured than the source datasets. Then we perform the segmentation using a state of the art segmentation model which is BiSeNet (Bilateral Segmentation Network) \cite{BiSeNet}. It is currently the fastest segmentation model tested on Cityscapes dataset\cite{cordts2016cityscapes} without affecting the accuracy. It achieves 74.7\% Mean IoU on the CityScape dataset with a speed of 65.5 frames per second\cite{cityscapes_dashboard}. The state of the art on CityScape dataset is PSPNet\cite{Zhao_2017} that achieves a Mean IoU 81.2\% but at very low speed: 0.78 frames per second\cite{cityscapes_dashboard}. The factor of speed is significantly important in aerial image processing as we need to process the video streams captured from aerial vehicles in real time. \(Figure\) \ref{fig:bisenet} represents the architecture of BiSeNet. 

The experiments related to this research work care concluded on a GPU machine with the following characteristics:\par
\begin{itemize}
\item CPU: Intel Core i9-8950HK (six cores, Coffee Lake architecture)
\item Graphic card: Nvidia GTX 1080, 8GB GDDR5
\item RAM: 32 GB RAM
\item Operating system: Linux (Ubuntu 16.04)
\end{itemize}
 
\begin{figure} [H]  
\begin{center}  
\includegraphics[width=8cm]{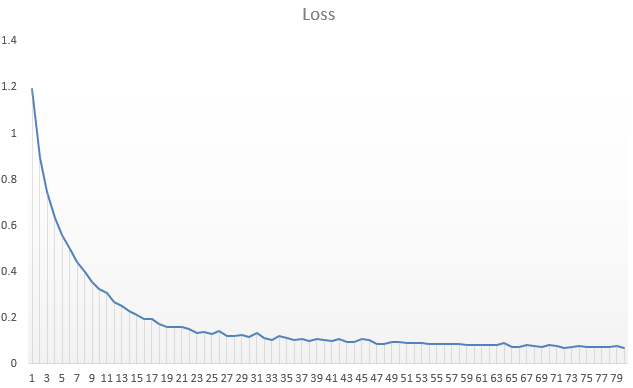}
\caption{\small \sl Loss of training BiseNet on Potsdam dataset\label{fig:loss_bisenet}}  
\end{center}  
\end{figure}

\begin{figure} [H]  
\begin{center}  
\includegraphics[width=8cm]{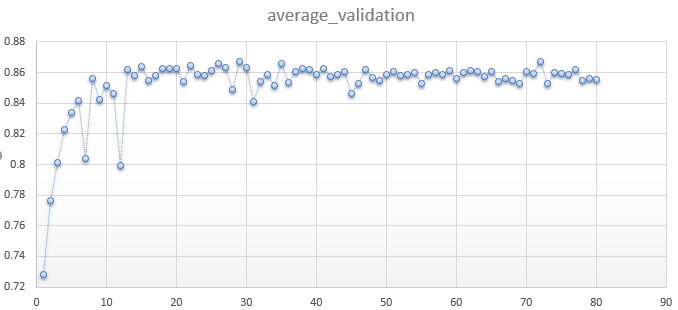}
\caption{\small \sl Evolution of average accuracy of BiseNet trained on Potsdam\label{fig:accuracy_bisenet_potsdam}}  
\end{center}  
\end{figure}

\begin{figure} [H]  
\begin{center}  
\includegraphics[width=8cm]{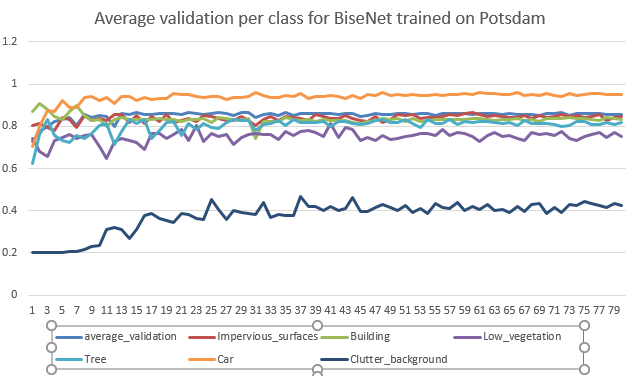}
\caption{\small \sl Evolution of per class accuracy of BiseNet trained on Potsdam \label{fig:Av_val_per_class_potsdam}}  
\end{center}  
\end{figure}

We used to train BiseNet on Potsdam the Semantic Segmentation Suite\cite{semantic_segmentation_suite}, which is an open source framework that provides the implementation of many segmentation models in Tensorflow\cite{tensorflow}. We used as feature extractor for BiSeNet, a state of the art network which is ResNet101\cite{ResNet}. We run the training for Postdam dataset for 80 epochs, batch size was 1 image per batch. We did not use image augmentation techniques. We used as optimizer for the training, ADAM optimizer\cite{kingma2014adam} with learning rate set to 0.0001. The training converges fast in less than 15 epochs and the average segmentation accuracy exceeds 86\%. \(Figure\) \ref{fig:loss_bisenet} shows the evolution of the training loss of BiSeNet on Potsdam dataset over epochs. 

 \(Figure\) \ref{fig:accuracy_bisenet_potsdam} shows the evolution of the segmentation accuracy of BiseNet on Potsdam validation dataset over epochs.  We can see that segmentation accuracy exceeds rapidly 86\% in few epochs. 
 \(Figure\) \ref{fig:Av_val_per_class_potsdam} shows the evolution of the segmentation accuracy of every class on Potsdam validation dataset over epochs.

After finishing the training, we saved the weights of BiSeNet model to be used later in Step 4 of the algorithm.
\subsubsection{Step2: Training our proposed GAN architecture}
To train our proposed GAN architecture described in section III.A, we constructed two datasets: one for Potsdam and the other for Vaihingen. For each dataset, We collected randomly 400 images of size \(512*512\) from  the original TOP images and divided these images into a training sub-set of 300 images and a test sub-set of 100 images. The proposed GAN architecture trains to translate images from Potsdam domain (source domain) to Vaihingen domain (target domain). 
The GAN-architecture is implemented using Keras\cite{chollet2015keras}, which is high level deep learning framework developed in Python. We used Tensorflow\cite{tensorflow} as a backend for the training. 
We set the slope \(\alpha\) for Leaky ReLU as 0.2. We used as optimizer for the training the ADAM optimizer\cite{kingma2014adam}, with a learning rate set to 0.0002. We trained the model until we got the discriminator accuracy superior to 92\% and and the generator loss inferior than 3. The convergence of the discriminator and the generator just need a few epochs of joint training. 

\subsubsection{Step3: Translating the source dataset to the target domain}
\begin{figure} 
\begin{center}  
\includegraphics[width=7cm]{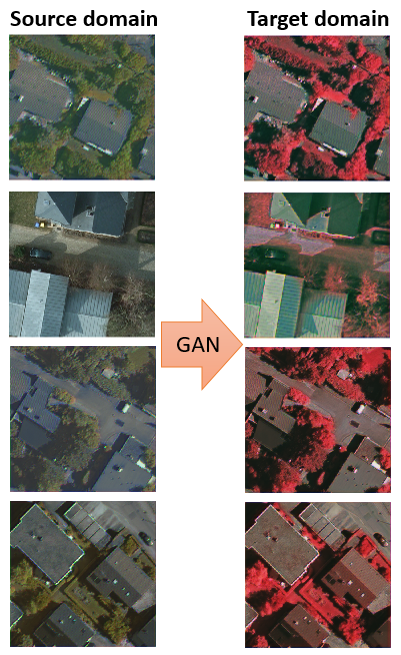}
\caption{\small \sl Mapping images from the source domain to the target domain using our proposed GAN \label{fig:mapping_samples}}  
\end{center}  
\end{figure}

Once the training of the proposed GAN architecture is done, we use it to translate the full dataset of the source domain (Potsdam) to the target domain (Vaihingen). \(Figure\) \ref{fig:mapping_samples} shows samples of Potsdam dataset translated to  Vaihingen domain. We note that the global style of the translated image is imitating the style of the target domain. The images generated are similar to what we can get as new images of Potsdam town using the IRRG sensor used for Vaihingen images. We keep this translated dataset to be used in the fourth step of our algorithm. 
\subsubsection{Step4: Fine-tuning the segmentation model with the translated dataset}
\begin{figure}[H]  
\begin{center}  
\includegraphics[width=7cm]{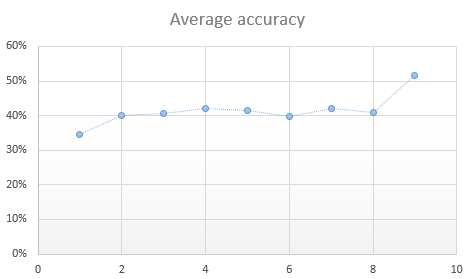}
\caption{\small \sl Improvement of average accuracy on Target dataset after each epoch
\label{fig:improvement_per_epoch}}  
\end{center}  
\end{figure}

\begin{figure} [H]  
\begin{center}  
\includegraphics[width=7cm]{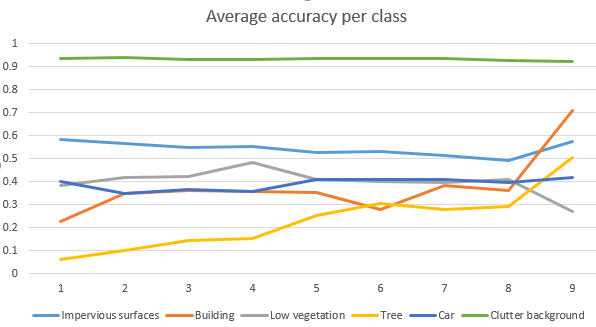}
\caption{\small \sl Improvement of accuracy per class on Target dataset after each epoch
\label{fig:improvement_per_class}}  
\end{center}  
\end{figure}

Once the translated dataset is ready, we use it to fine-tune the trained model prepared on Step 1. We do the fine-tuning process epoch by epoch and we tested the model on the target dataset after every epoch to measure the improvement of average accuracy on the target dataset. We noted an increase in average accuracy between 5\%and 17\%. Average accuracy value is improved from 34\% to values between 39\% and 52\%. We got an increase of 17\% after 8 epochs only. In \(Figure\) \ref{fig:improvement_per_epoch}, we draw the improvement in average accuracy on the target dataset (Vaihingen) after every epoch of the fine-tuning process. 
In \(Figure\) \ref{fig:improvement_per_class} we show the improvement of per class accuracy on the target dataset (Vaihingen) after every epoch of the fine-tuning process. We can see in the figure that the accuracy of two classes (Trees and Building) increases highly over epochs although the other remains practically the same. In Section 6, we discuss the obtained results and we demonstrate the utility of our proposed approach in the domain adaptation of aerial imagery

\section{Discussion}
\begin{figure*}  
\begin{center}  
\includegraphics[width=15cm]{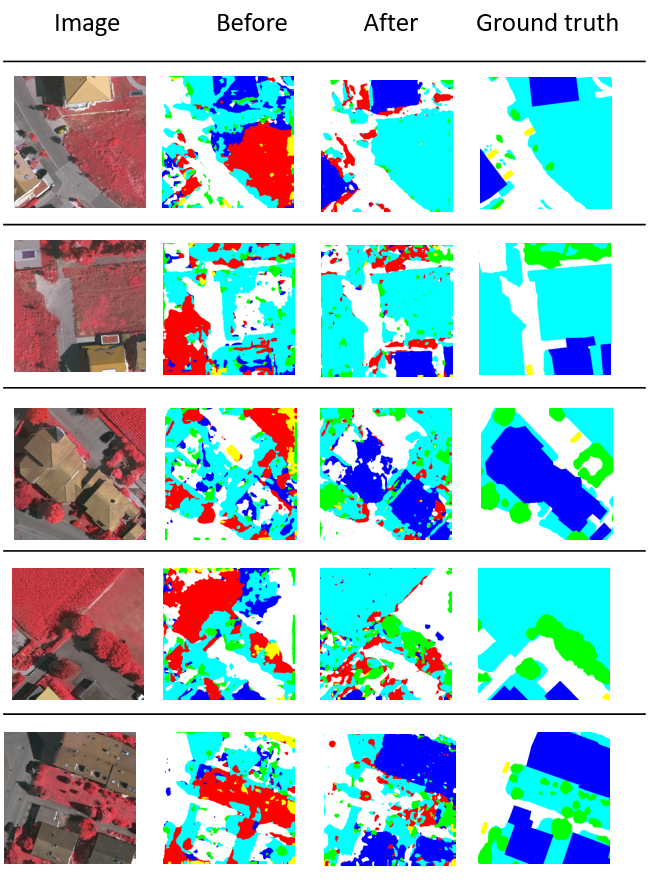}
\caption{\small \sl Samples of segmentation before and after the implementation of our algorithm
\label{fig:samples_improvements}}  
\end{center}  
\end{figure*}

The implementation of our algorithm increases the average accuracy of the segmentation model on the target dataset by a significant margin that reaches 17\%. Also, as presented in \(Table\) \ref{tab_general_results}, similar improvements are seen also in the precision, recall, F1 and IoU (Intersection over Union) measures. These improvements  made a visible amelioration on the predicted segmentation mask as presented in the \(Figure\) \ref{fig:samples_improvements}.
\begin{table} [H]
\centering
\begin{tabular}{lll}
\hline
 & \textbf{Before}  & \textbf{After}  \\ \hline
\textbf{Average accuracy} & \textbf{0.35} & \textbf{0.52} \\ 
\textbf{Precision} & \textbf{0.35} & \textbf{0.54} \\ 
\textbf{Recall} & \textbf{0.35} & \textbf{0.52} \\ 
\textbf{F1 measure} & \textbf{0.32} & \textbf{0.49} \\ 
\textbf{IoU score} & \textbf{0.17} & \textbf{0.30} \\ \hline
\end{tabular}
\caption{Segmentation metrics on the target dataset before and after the implementation of our algorithm}
\label{tab_general_results}
\end{table}

Going deeper, we made a study of the effect of our algorithm on every class apart. As described in the \( Table\) \ref{per_class_results}, we have two  types of effects. First, we have classes that our algorithms had increased the model accuracy by a high margin (classes Building and Tree). Comparing these results with \(Table\) \ref{tab:domain_shift_factors}, we note that these classes are characterized by a domain shift related highly to the sensor factor. If the domain shift is related only to the sensor factor, our algorithm is very efficient in increasing the accuracy of the model. For example, the class Building, as explained in IV-A-2, is only affected by the sensor factor. We can see its average accuracy increasing from 0.23 to 0.71. If the domain shift is related mostly to the sensor factor like class Tree, our algorithm will be very efficient in increasing the accuracy but with some limitations due to the other domain shift factors. Concerning other classes (Impervious surfaces, Car, Clutter background and Low vegetation), we can note that our algorithm has no practical effect in increasing or decreasing the accuracy. Accuracy will be conserved by our algorithm. These classes are, as described in  \(Table\) \ref{tab:domain_shift_factors}, either not affected by any domain shift factor (like classes Cars and Impervious surfaces ) or highly affected by other factor than sensor factor (like Clutter Background or Low vegetation).

\begin{table}[H]
\centering
\begin{tabular}{lll}
\hline
 & {\ul \textbf{Before}} & {\ul \textbf{After}} \\ \hline
 {\ul \textbf{Building}} & {\ul \textbf{0.23}} & {\ul \textbf{0.71}} \\
 {\ul \textbf{Tree}} & {\ul \textbf{0.06}} & {\ul \textbf{0.51}} \\ \hline
Impervious surfaces & 0.58 & 0.57 \\
Car & 0.40 & 0.42 \\
Clutter background & 0.94 & 0.93 \\ 
Low vegetation & 0.38 & 0.27 \\ \hline
\end{tabular}
\caption{Accuracy of the segmentation on every class before and after the implementation of our algorithm}
\label{per_class_results}
\end{table}

 We can estimate, that our algorithm conserves the accuracy of the model if there is no domain shift or if the domain shift is related highly to other factor than the sensor factor. This is a highly appreciated feature as it allows combining it with other techniques that may reduce other domain shift factors. Our algorithm targets successfully the elimination of the sensor  factor without affecting other factors. If the domain shift between the source dataset and the target dataset is only related to it, our algorithm is capable of improving the accuracy to a level similar to training the model on a full labeled dataset of the target, as seen in the class Building. This fact is very helpful for aerial imagery processing, as it will relieve us from making new labeling dataset. \(Table\) \ref{per_class_results} resumes the efficiency of our algorithm per case.

\begin{table}[H]
\centering
\begin{tabular}{lll}
\hline
\textbf{Domain shift factor} & \textbf{Efficiency} & \textbf{Examples of classes} \\ \hline
\textbf{Sensor} & \textbf{High}  & Building, Tree \\ \hline
\textbf{Other factors} & \textbf{Conserve efficiency}   & Low vegetation \\ \hline
\textbf{No Domain shift} & \textbf{Conserve efficiency} & Cars \\ \hline
\end{tabular}
\caption{Efficiency of our algorithm per case}
\label{per_domain_shift_factor}
\end{table}

\section{Conclusion}
In this work, we have proposed a new method for domain adaptation in semantic segmentation in aerial imagery based on GANs. This method confirm to be efficient in targeting domain shift that results from  sensor variation between the source and the target. In this case, this method is capable to improve substantially the accuracy of the segmentation model. Besides, it does not affect the ability of the segmentation model to classify classes that are not have domain shift or classes that are subject for other domain shift factors like variation of resolution or variation of class representation. Moreover, it has a very minimal cost as it does not need labeling data or other manual work. It just requires building dataset from random images collected arbitrarily from the source dataset and the target dataset. After, we have to apply automatically the four steps of the method to enhance significantly the ability of the model in domain adaptation between the source and the target.
In a future development, we would like to extend this method to handle semi-supervised domain scenarios where a few labels are available in the target domain.



\authorcontributions{Bilel Benjdira and Yakoub Bazi designed the method. Bilel Benjdira implemented the method and wrote the paper. Yakoub Bazi, Anis Koubaa and Kais Ouni contributed to the supervision of the work, analysis of the method and paper writing.}

\funding{This research was funded by Prince Sultan University.}

\acknowledgments{This work is supported by Robotics and Internet of Things Lab (RIOTU), Prince Sultan University.}

\conflictsofinterest{The authors declare no conflict of interest.}

\reftitle{References}

\externalbibliography{yes}
\bibliography{biblio}
\end{document}